\pdfoutput=1

\documentclass[11pt,a4paper]{article}
\usepackage[hyperref]{eacl2021}
\usepackage{url}

\usepackage{breakurl}
\aclfinalcopy

\usepackage{times}
\usepackage{CJKutf8}
\usepackage{amssymb}
\usepackage{amsmath}
\usepackage{mathtools}
\usepackage{bm}
\usepackage{textcomp}
\usepackage{varwidth}
\usepackage{caption}
\usepackage{enumitem}
\usepackage{placeins}
\usepackage{times}
\usepackage{latexsym}
\usepackage{multirow}
\usepackage{multicol}
\usepackage{makecell}
\usepackage{arydshln}
\usepackage{threeparttable}
\usepackage{subcaption}
\usepackage{textgreek}

\title{Approaching Neural Chinese Word Segmentation\\ as a Low-Resource Machine Translation Task}

\author{Pinzhen Chen \qquad\qquad Kenneth Heafield\\
    School of Informatics, University of Edinburgh \\
 \texttt{\{pinzhen.chen, kenneth.heafield\}@ed.ac.uk}}

\date{}

\begin{document}
\maketitle
\begin{abstract}
Chinese word segmentation has entered the deep learning era which greatly reduces the hassle of feature engineering. Recently, some researchers attempted to treat it as character-level translation, which further simplified model designing, but there is a performance gap between the translation-based approach and other methods. This motivates our work, in which we apply the best practices from low-resource neural machine translation to supervised Chinese segmentation. We examine a series of techniques including regularization, data augmentation, objective weighting, transfer learning, and ensembling. Compared to previous works, our low-resource translation-based method maintains the effortless model design, yet achieves the same result as state of the art in the constrained evaluation without using additional data.
\end{abstract}

\section{Introduction}
Chinese text is written in characters as the smallest unit, and it has no explicit word boundary. Therefore, Chinese word segmentation (CWS) serves as upstream tokenization and disambiguation for Chinese language processing. The task is often viewed as sequence labelling, where each character receives a label indicating its relative position in a segmented sequence (e.g. whether the character is at the word boundary). While traditional machine learning methods have attained strong results, recent investigations focus on neural networks given their rise in the entire NLP field. Distinctive to sequence labelling, \citet{shi-etal-2017-neural} first treat CWS as neural machine translation (NMT). Nonetheless, \citet{zhao-etal-2018-chinese} point out that without extra resources, all previous neural methods are not yet comparable with the non-neural state of the art from \citet{zhao-kit-2008-unsupervised}; the NMT practice is even behind.

We note two advantages of treating the task as neural translation: 1) the entire input sentence is encoded before making any segmentation decision; 2) such a model jointly trains character embeddings with sequence modelling. Thus, we try to bridge the gap between the translation-based approach and state-of-the-art models, using low-resource techniques commonly seen in NMT. The translation-based method can be easy to adopt without the need for feature extraction and model modifictaion. Although NMT is known to be data hungry, our approach is able to achieve competitive results in the constrained evaluation scenario, where introducing extra data is forbidden. In specific, when benchmarked on the second CWS bakeoff \citep{emerson-2005-second}, our system reaches the top of the MSR leaderboard and achieves a strong result on the PKU dataset.

\section{Related Work}
Chinese segmentation is traditionally tackled as sequence labelling, which predicts whether each input character should be split from neighbouring characters \citep{xue-2003-chinese}. Earlier approaches relied on conditional random fields or maximum entropy Markov models \citep{peng-etal-2004-chinese, ng-low-2004-chinese}. \citet{zhao-kit-2008-unsupervised} leveraged unsupervised features to attain state-of-the-art results in the data-constrained track. 

Recent research has shifted towards neural networks: feed-forward, recurrent and convolutional \citep{zheng-etal-2013-deep, pei-etal-2014-max, chen-etal-2015-gated, chen-etal-2015-long, wang-xu-2017-convolutional}. Without external data, these models did not surpass the best non-neural method, but instead, provided great ease of data engineering. Researchers also studied better representations for segments and characters, as well as the incorporation of external resources \citep{liu-etal-2016-exploring-segment, zhou-etal-2017-word, yang-etal-2017-neural-word}. By carefully tuning model configurations, \citet{ma-etal-2018-state} achieved strong results. The task can also be done through learning to score global word segmentation schemes given characters \citep{zhang-clark-2007-chinese, zhang-clark-2011-syntactic, cai-zhao-2016-neural, cai-etal-2017-fast}. On top of this, \citet{wang-etal-2019-unsupervised-learning} proved that it is beneficial to integrate unsupervised segmentation. A recent work used a modified Transformer for sequence tagging to attain the same results as the state of the art \citep{duan-zhao-2020-attention}.

The most relevant to our research is \citet{shi-etal-2017-neural}'s suggestion to formalize Chinese word segmentation as character-level neural machine translation. It differs from global segmentation scoring in that the NMT directly generates Chinese characters with delimiters. It can also be equipped with post-editing that adds back characters omitted by the model. Later, \citet{wei-etal-2019-improved} restrict the NMT decoding to follow all and only the input characters. This proposal, together with existing NMT toolkits, eases the model design and implementation for neural Chinese segmentation. However, even with external resources, the two systems are inferior to the previous works concerning performance. This encourages us to explore low-resource techniques to enhance the NMT-based approach.

\section{Methodology}
An NMT model is trained to minimize the sum of an objective function $L$ over each target sentence $y_{0}^{n}=y_0, y_1,...,y_n$ given a source sentence $X$. We list below per-character conditional cross-entropy as an example:
\begin{align}
%\label{eq:objective-func}
   L = - \frac{1}{n} \sum \limits_{i=1}^n \log P(y_i|y_{0}^{i-1},X)
\end{align}
Following \citet{shi-etal-2017-neural}, we make use of character-level NMT, and add an extra delimiter token ``$\langle D\rangle$'' to the target vocabulary. The delimiter token on the target side implies that the previous and next words are separated. To visualize, given an unsegmented sentence in characters ``\begin{CJK*}{UTF8}{gbsn}我 会 游 泳\end{CJK*}'', the model should output character-by-character ``\begin{CJK*}{UTF8}{gbsn}我 $\langle D\rangle$ 会 $\langle D\rangle$ 游 泳\end{CJK*}'' (English: I can swim). This in reality resembles how a human would read an unsegmented Chinese sentence.

We argue that NMT can model word segmentation well because the decoder has access to the global information in both decoder and attention states. Moreover, the output segmented characters may display stronger probabilistic patterns than the position labels do, resulting in more explicit modelling of the word boundary ``$\langle D\rangle$''. This characteristic is also robust to out-of-vocabulary words because NMT can freely ``insert'' the boundary delimiter anywhere to form words. Finally, this method does not require any alteration to the model architecture.

However, it poses a challenge when CWS is approached as NMT, that NMT usually performs poorly under a low-resource condition \citep{koehn-knowles-2017-six}, which is exactly the case of Chinese segmentation datasets. A CWS corpus provides fewer than 100k sentences, whereas a typical translation task provides data at least one order of magnitude larger. To address this issue, we apply low-resource NMT practices: regularization and data augmentation. Then, we examine several other broadly used techniques.

\subsection{Hyperparameter tuning}
Hyperparameter tuning is often the first step to build a machine learning model. In the field of neural translation, \citet{sennrich-zhang-2019-revisiting} show that carefully tuning hyperparameters results in substantial improvement in low-resource scenarios. In our case, we concentrate on regularization: label smoothing, network dropout, and source token dropout \citep{szegedy-etal-2016-rethinking, srivastava-etal-2014-dropout, sennrich-etal-2016-edinburgh}. Additionally, we switch between GRU and LSTM, and increase the model depth \citep{hochreiter-schmidhuber-1997-long, cho-etal-2014-learning}.

\subsection{Objective weighting}
The generic NMT objective function considers the loss from each target sentence or token equally. By adjusting the objective function we can make it weigh some components more than others, in order to better learn the desired part of the training data. It can be applied at the token or sentence level, for various purposes including domain adaptation and grammatical error correction \citep{chen-etal-2017-cost, wang-etal-2017-instance, yan-etal-2018-word, junczys-dowmunt-etal-2018-approaching}. 

We propose to put more emphasis on the delimiter token in target sentences because they correspond to word boundaries directly. We weight delimiters $k$ times as many as other tokens in the objective function, where $k$ can be empirically determined on a validation set. The new token-weighted objective function $L_{token}$ is as Equation~\ref{eq:token-weighted}, where the weight coefficient $\bm{\lambda}_i = k$ if $y_i$ is a delimiter and $\bm{\lambda}_i = 1$ otherwise.  
\begin{equation}
\label{eq:token-weighted}
L_{token} = - \frac{1}{n} \sum \limits_{i=1}^n \bm{\lambda_{i}} \log P(y_i|y_{0}^{i-1},X) %
\end{equation}

\subsection{Data augmentation}
Data augmentation is widely adopted in NMT. The paradigm is to generate source side data from existing (monolingual) target side data \citep{sennrich-etal-2016-improving, grundkiewicz-etal-2019-neural}, but this does not apply to CWS since there is no extra gold segmented data. Hence, we experiment with two methods that could suit CWS better: sentence splitting and unsupervised segmentation.

\subsubsection{Sentence splitting}
The surface texts of inputs and outputs are consistent in the NMT approach to CWS, with the only exception being the added delimiters. With a potential quality degrade, we assume that segmentation can be inferred locally, i.e.\ within a phrase instead of the whole sentence. It enables us to split a sentence into multiple shorter segments, with the gold segmentation unchanged. This can hugely expand the amount of training data, and reduce the input and output sequence lengths. In practice, we break full sentences down at comma and period symbols, since they are always separated from other characters.

\subsubsection{Unsupervised segmentation}
\label{sec:unsup}
Both \citet{zhao-kit-2008-unsupervised}'s, and \citet{wang-etal-2019-unsupervised-learning}'s papers show that unsupervised segmentation helps supervised CWS. We use an external tool to segment our training data in an unsupervised way, to create augmented data (detailed later in Section~\ref{sec:unsup-segmenter}). The data is utilized in two scenarios different from previous works: sentence-level weighting and transfer learning. The methods are depicted below.

\paragraph{Sentence weighting} Weighting objective function at the sentence level can distinguish high- and low-quality training data. We designate our unsupervised segmentation result as low-quality augmented data, and the original training sentences as high-quality data. After combining them into a single training set, the high-quality data is weighted $k$ times as much as the low-quality data. Equation~\ref{eq:sent-weighted} shows that sentence-weighted objective function $L_{sentence}$, where weight $\bm{\lambda} = k$ for gold sentences and $1$ for augmented sentences. In contrast with Equation~\ref{eq:token-weighted}, the sentence weight is not token-dependant.
\begin{equation}
\label{eq:sent-weighted}
   L_{sentence} = - \frac{1}{n} \sum \limits_{i=1}^n \bm{\lambda} \log P(y_i|y_{0}^{i-1},X)
\end{equation}

\paragraph{Transfer learning} It means to pre-train a model on high-resource data and then optimize it for a low-resource task. It often yields enhanced results over directly training on a small dataset \citep{zoph-etal-2016-transfer}, as the knowledge learned from the high-resource task can be beneficial. Moreover, \citet{aji-etal-2020-neural} claim that starting a model from trained parameters is better than random initialization. We first train a model on the augmented data from an unsupervised segmenter, then further optimize it on the genuine training data.

\subsection{Ensembling}
An ensemble of diverse and independently trained and models enhances prediction. In our work, we combine models trained with different techniques and random seeds, and integrate a neural generative language model (LM) trained on the gold segmented training data. It works as follows: at each inference time step, all models' predictions are simply averaged to form the ensemble's prediction over the target vocabulary.

\section{Experiments and Results}
\subsection{Task description}
Evaluation takes place on the Microsoft Research (MSR) and Peking University (PKU) corpora in the second CWS bakeoff \citep{emerson-2005-second}.\footnote{\href{http://sighan.cs.uchicago.edu/bakeoff2005/}{sighan.cs.uchicago.edu/bakeoff2005}.} The datasets are of sizes 87k and 19k, which are considered low-resource in machine translation tasks. Regarding preprocessing, our own training and validation sets are created randomly at a ratio of 99:1, from the supplied training data. We normalize characters, and convert continuous digits and Latin alphabets to ``$\langle N\rangle$'' and ``$\langle L\rangle$'' symbols without affecting segmentation.

There are both closed (constrained) and open tests in the CWS bakeoff. The former requires a system to only use the supplied data. Since we aim to strengthen the translation-based approach itself, we select the closed test condition and compare with other papers that report closed test results. The evaluation metric F1 (\%) is calculated by the script from the bakeoff. We test different techniques on MSR and apply the best configurations to PKU without further tuning. 

\begin{table}[tbh]
\begin{subtable}[htb]{\linewidth}
\centering
    \begin{tabular}{|c|c|c|c}
        \cline{1-3}
        & drop\textsubscript{state} & best loss \\
        \cline{1-3}
       \multirow{5}{*}{\makecell{drop\textsubscript{src}\\= 0}} 
        & 0            & 0.0333 \\
        & 0.1 & 0.0271 \\
        & 0.2 & \textbf{0.0262} & \checkmark \\
        & 0.3 & 0.0272 \\
        & 0.4          & 0.0303 \\
        \cline{1-3}
        \multicolumn{3}{c}{} \\
        \cline{1-3}
        & drop\textsubscript{src} & best loss \\
        \cline{1-3}
        \multirow{3}{*}{\makecell{drop\textsubscript{state}\\= 0.2}} 
        & \textbf{0} & \textbf{0.0262} & \checkmark \\
        & 0.15       & 0.2081 \\
        & 0.3        & 0.4496 \\
        \cline{1-3}
    \end{tabular}
\caption{Experiments on two dropout methods. drop\textsubscript{src} indicates entire source word dropout and drop\textsubscript{state} indicates dropout between RNN states.}
\label{tab:dropout}
\vspace{1ex}
\end{subtable}

\begin{subtable}[tbh]{\linewidth}
\centering    
    \begin{tabular}{|c|c|c|c}
        \cline{1-3} 
        & label smoothing & best loss \\
        \cline{1-3} 
       \multirow{3}{*}{\makecell{drop\textsubscript{src}= 0,\\drop\textsubscript{cell}= 0.2}} 
        & \textbf{0} & \textbf{0.0262} & \checkmark \\
        & 0.1        & 0.1161 \\ 
        & 0.2        & 0.2220 \\ 
        \cline{1-3} 
    \end{tabular}
\caption{Experiments on label smoothing.}
\label{tab:label-smoothing}
\vspace{1ex}
\end{subtable}

\begin{subtable}[tbh]{\linewidth}
\centering    
\begin{tabular}{|c|c|c|c|c}
    \cline{1-4}
    cell &
    \makecell{encoder\\depth} & 
    \makecell{decoder\\depth} & 
    best loss \\
    \cline{1-4}
    \multirow{6}{*}{GRU} 
      & 1 & 1 & 0.0262 & \checkmark \\
      & 1 & 2 & 0.0251 \\
      & 2 & 1 & 0.0261 \\
      & 2 & 2 & 0.0264 \\
      & 3 & 3 & 0.0276 \\
      & 4 & 4 & 0.0268 \\
    \cdashline{1-4}
    LSTM & 1 & 1 & 0.0286 \\
    \cline{1-4}
\end{tabular}
\caption{Experiments on model depth and the RNN type. No obvious winner is observed.}
\label{tab:num-of-layers}
\end{subtable}
\caption{Hyperparameter searches.}
\label{tab:hyperparameter-searches}
\end{table}

\subsection{Baseline with regularization}
\label{sec:param-tuning}
We start with a 1-layer bi-directional GRU with attention \citep{bahdanau-etal-2015-neural} containing 36M parameters. Adam \citep{kingma-ba-2015-adam} is used to optimize for per-character (token) cross-entropy until the cost on the validation set stalls for 10 consecutive times. We set the learning rate to $10^{-4}$, beam size to 6, and enable layer normalization \citep{ba-etal-2016-layer}. Since the model input and output share the same set of characters, we use a shared vocabulary and tied embeddings for source, target, and output layers \citep{press-wolf-2017-using}. Training such a model on the MSR dataset takes 5 hours on a single GeForce GTX TITAN X GPU with the Marian toolkit \citep{junczys-dowmunt-etal-2018-marian-fast}.\footnote{\href{https://github.com/marian-nmt/marian}{https://github.com/marian-nmt/marian}.}

Regarding hyperparameter selection, we always select the best settings based on the loss on the validation set. The tuning procedures are reported in Table~\ref{tab:hyperparameter-searches}. We see that a small dropout of 0.2 is helpful; source token dropout and label smoothing both cause adverse effects. Changing model depth and switching from GRU to LSTM make a negligible impact, so we stick to the single-layer GRU.

The first row in Table~\ref{tab:internal-results} shows that our carefully-tuned baseline achieves an F1 of 96.8\% on the MSR test set. Next, we find that weighting delimiters twice as other tokens brings a 0.1\% improvement. Delimiter weight tuning is presented in Table~\ref{tab:word-weighting}. These numbers already outperform previous translation-based works.

\begin{table}[bht]
\centering    
    \begin{tabular}{|c|c|c}
        \cline{1-2}
        \makecell{weight $\lambda$\\on delimiters} & best loss \\
        \cline{1-2}
        1 (no weighting) & 0.0262 \\
        \cdashline{1-2}
        1.5 & 0.0197 \\
        \textbf{2} & \textbf{0.0191} & \checkmark \\
        4 & 0.0204 \\
        10 & 0.0210 \\
        50 & 0.0253 \\
        \cline{1-2}
    \end{tabular}
\caption{Experiments on delimiter (word) weighting. $\lambda$ is the weight on the delimiter, and other words are always given a weight of 1.}
\label{tab:word-weighting}
\end{table}

\begin{table}[htb]
\centering    
    \begin{tabular}{|c|c|c}
        \cline{1-2}
        \makecell{weight $\lambda$ on\\original data} & best loss \\
        \cline{1-2}
        1 (no weighting) & 0.0462 \\
        \cdashline{1-2}
        2 & 0.0346 \\
        5 & 0.0309\\
        10 & 0.0268\\
        20 & 0.0227 \\
        \textbf{40} & \textbf{0.0226} & \checkmark\\
        100 & 0.0230 \\
        200 & 0.0245 \\
        \cdashline{1-2}
        only genuine data  & 0.0268 \\
        \cline{1-2}
    \end{tabular}
\caption{Experiments on weighting augmented and original data. $\lambda$ represents the weight on original sentences; augmented data always have a weight of 1.}
\label{tab:sentence-weighting}
\end{table}

\begin{table}[bht]
\centering
\begin{tabular}{|l|c|}
\hline
    \multicolumn{1}{|c|}{Techniques} & F1 (\%) \\
\hline
% (w/ hyper-parameter tuning)
    baseline w/ regularization (base) & 96.8 \\
\hdashline
    base + delimiter weight & 96.9 \\
    base + sentence splitting (split) & 97.1 \\
    base + split + unsupervised + transfer & 97.1 \\
    base + split + unsupervised + weight & \textbf{97.3} \\
\hdashline
    2 $\times$ baseline & 97.2 \\
    2 $\times$ transfer + 2 $\times$ weight + LM & \textbf{97.6} \\
\hline
\end{tabular}
\caption{F1 of our techniques on MSR test set.}
\label{tab:internal-results}
\end{table}

\subsection{Leveraging augmented data}
Sentence splitting is done on both sides of the training and validation sets. Test sentences are split, segmented by the model, and then concatenated, ensuring a consistent evaluation outcome. This leads to a better F1 of 97.1\%, thanks to a 3-fold increase in data size to 257k for MSR.

\label{sec:unsup-segmenter}
We employ the segmental language model \citep{sun-deng-2018-unsupervised} for unsupervised segmentation.\footnote{Their code and released models: \href{https://github.com/edward-sun/slm}{github.com/edward-sun/slm}.} We used the MSR model optimized on the training, validation, and test sets with a maximum word length of 4. Since the system is fully unsupervised, it is fair to include the test set; yet we only apply it to our training split to generate augmented data. In this way, no external resource is introduced. While transfer learning brings no gain, sentence-level weighting lifts the overall score to 97.3\%, as shown in Table~\ref{tab:internal-results}. We see that the cost on the validation set improves, and then degrades as sentence weight gets larger; the best sentence weight is determined to be 40 for MSR. The detailed weight selection process is described in Table~\ref{tab:sentence-weighting}.

\subsection{Ensembling}
During decoding, all models' predictions are averaged to produce an output token at each step. We first test an ensemble consisting of two baselines. Next, we combine two transfer learning models, two sentence-weighting models, and a character RNN LM. The LM has the same architecture as our NMT decoder. It is optimized for perplexity on the segmented side of the train set. Ensembling is done in one shot without tuning weights and it achieves the highest F1 of 97.6\%.

\section{Results and Analysis}
In addition to MSR test, we keep the best hyperparameters determined on the MSR corpus unchanged, and run the same set of experiments on the PKU dataset. 

\begin{table}[htb]
\centering
\begin{threeparttable}[htb]
\begin{tabular}{|c|l|c|c|}
\hline
    \multicolumn{2}{|c|}{System} & MSR & PKU \\
\hline
    \multirow{2}{*}{\makecell{non-\\neural}} &
      \citealp{zhao-kit-2008-unsupervised} & \textbf{97.6} & 95.4 \\
    & \citealp{zhang-clark-2011-syntactic} & 97.3 & 94.4 \\
\hline
    \multirow{8}{*}{\makecell{neural}} & 
    % &  \citealp{zheng-etal-2013-deep}\tnote{\P} & 93.3 \\
     \citealp{pei-etal-2014-max} & 94.4 & 93.5 \\
    % & \citealp{chen-etal-2015-gated}\tnote{\S} & 95.1 \\
    % & \citealp{chen-etal-2015-long}\tnote{\S} & 95.0 \\
    & \citealp{cai-zhao-2016-neural} & 96.4 & 95.2 \\
    & \citealp{wang-xu-2017-convolutional} & 96.7 & 94.7 \\
    & \citealp{cai-etal-2017-fast} & 97.0 & 95.4 \\
    & \citealp{zhou-etal-2017-word} & 97.2 & 95.0 \\
    & \citealp{ma-etal-2018-state} & \textbf{97.5} & 95.4 \\
    & \citealp{wang-etal-2019-unsupervised-learning} & 97.4 & \textbf{95.7} \\
    & \citealp{duan-zhao-2020-attention} & \textbf{97.6} & 95.5 \\
\hline
     \multirow{5}{*}{\makecell{NMT-\\based}} & 
      \citealp{shi-etal-2017-neural} & 94.1 & 87.8 \\
    & \hspace{6pt} + external resources\tnote{\dag} & 96.2 & 95.0 \\
    & \citealp{wei-etal-2019-improved}\tnote{\dag} & 94.4 & 92.0 \\
\cdashline{2-4}
    & Our best single model & 97.3 & 95.0 \\
    & Our best ensemble\tnote{\ddag} & \textbf{97.6} & 95.4 \\
\hline
\end{tabular}
\begin{tablenotes}
\footnotesize
     \item[\dag] The results are advantaged as extra resources are used.
     \item[\ddag] 97.61$\pm$0.16 on MSR and 95.43$\pm$0.38 on PKU, with $p<0.05$ using bootstrapping \citep{ma-etal-2018-state}, detailed in Appendix~\ref{sec:significance-testing}.
   \end{tablenotes}
\end{threeparttable}
\caption{Previous and our systems' F1 (\%) on MSR and PKU corpora under the constrained condition.}\label{tab:results}
\end{table}

Table~\ref{tab:results} compares our MSR and PKU results with previous papers. Our best single models are remarkably ahead of other NMT-based methods. With ensembling, our result on MSR ties with state of the art, showing that empirically neural methods can reach the top without external data. However, as data size significantly drops in the case of PKU, we observe a declined performance and larger variance on the PKU dataset. This is expected as NMT is known to be sensitive to a smaller data size.

Regarding regularization, we discover that low-resource NMT techniques are not always constructive for CWS. Dropping out source tokens is harmful because CWS is not a language generation task and the decoder output heavily relies on the input. A similar rationale explains why label smoothing causes rocketing cross-entropy: unlike language generation where a variety of outputs are accepted, for CWS there is always just one single correct scheme. Smoothing out the decoder probability distribution results in confusion.

Further, unsupervised data augmentation with weighting achieves the best single-model result. We suggest a possible reason: the augmented data has the same source side as the original data, but a noisier target side. When weighted appropriately, the noise might act as a smoothing technique for sequence modelling, especially in the low-resource condition 
\citep{xie-etal-2017-data}. From the transfer learning aspect, pre-training on the augmented data does not lead to a higher number than starting from a randomly initialized state.

\section{Conclusion}
Our low-resource translation-based approach to Chinese word segmentation achieves strong performance and is easy to adopt. Data augmentation, objective weighting and ensembling are the most favourable. In future, it is worth extending this perspective to word segmentation of other languages, as well as re-basing it on Transformer models.

\section*{Acknowledgements}
This research is supported in part by the Office of the Director of National Intelligence (ODNI), Intelligence Advanced Research Projects Activity (IARPA), via contract \#FA8650-17-C-9117. The work reflects the view of the authors and not necessarily that of the funders.

\bibliography{eacl2021}
\bibliographystyle{acl_natbib}

\vspace{2ex}
\appendix
\section{Results with a confidence interval}
\label{sec:significance-testing}
We report our final score with a confidence interval since the top results are very close. As there is only one test set, we create another 599 test sets of the same size as the original one, through resampling with replacement. Our best system obtains an F1 of 97.61$\pm$0.16 on the MSR dataset and  95.43$\pm$0.38 on the PKU dataset with 95\% confidence (2 standard deviations).

\end{document}